%% file: main.tex
\documentclass{article}
\usepackage{ijcai19}

\usepackage{times}
\usepackage{soul}
\usepackage{url}
\usepackage[hidelinks]{hyperref}
\usepackage[utf8]{inputenc}
\usepackage[small]{caption}
\usepackage{graphicx}
\usepackage{amsmath}
\usepackage{booktabs}
\usepackage{subcaption}
\usepackage{float}
\usepackage{bm}

\usepackage{soul}
\usepackage{color}

\usepackage{amssymb}
\DeclareMathOperator*{\argmax}{arg\,max}
\DeclareMathOperator*{\argmin}{arg\,min}
\newcommand{\counterfactual}{\ensuremath{%
  \Box\kern-1.5pt
  \raise1pt\hbox{$\mathord{\rightarrow}$}}}
\newcommand{\briefCite}[1]{\citeauthor{#1}(\citeyear{#1})}

\title{Counterfactual States for Atari Agents via Generative Deep Learning}
\author{Matthew L. Olson
\and
Lawrence Neal\and
Fuxin Li \and
Weng-Keen Wong
\affiliations
Oregon State University
\emails
\{olsomatt, nealla, lif, wongwe\}@eecs.oregonstate.edu
}

\begin{document}
\maketitle
\begin{abstract}
Although deep reinforcement learning agents have produced impressive results in many domains, their decision making is difficult to explain to humans. To address this problem, past work has mainly focused on explaining why an action was chosen in a given state. A different type of explanation that is useful is a counterfactual, which deals with ``what if?'' scenarios. In this work, we introduce the concept of a ${\it counterfactual}$ ${\it state}$ to help humans gain a better understanding of what would need to change (minimally) in an Atari game image for the agent to choose a different action. We introduce a novel method to create counterfactual states from a generative deep learning architecture. In addition, we evaluate the effectiveness of counterfactual states on human participants who are not machine learning experts. Our user study results suggest that our generated counterfactual states are useful in helping non-expert participants gain a better understanding of an agent's decision making process.

\end{abstract}

\section{Introduction}
\input{introduction.tex}

\section{Related Work}
\input{relatedwork.tex}

\section{Methodology}
\input{methodology.tex}

\section{Results}
\input{experiments.tex}

\section{Discussion}

\input{discussion.tex}

\section{Conclusion}
\input{conclusion.tex}

\clearpage
\bibliography{ijcai19}
\bibliographystyle{named}


\end{document}

%% file: introduction.tex


Although deep reinforcement learning (RL) agents have produced impressive results, their decision-making process is often inscrutable to humans. This limitation is a serious roadblock for applications in which trust and reliability are critical. In order to solve this problem, researchers have begun developing techniques to peer inside these ``black boxes''. The majority of these techniques provide explanations as to why the agent chose a particular action (e.g. \cite{pmlr-v80-greydanus18a}). We present a different, but complementary, type of explanation based on counterfactuals \cite{lewis2013counterfactuals}, which deal with ``what if?'' scenarios. Specifically, a counterfactual explanation describes what would need to change in order for the agent to choose a different action. 

\begin{figure}[t]
    \centering
    {{\includegraphics[width=3.845cm]{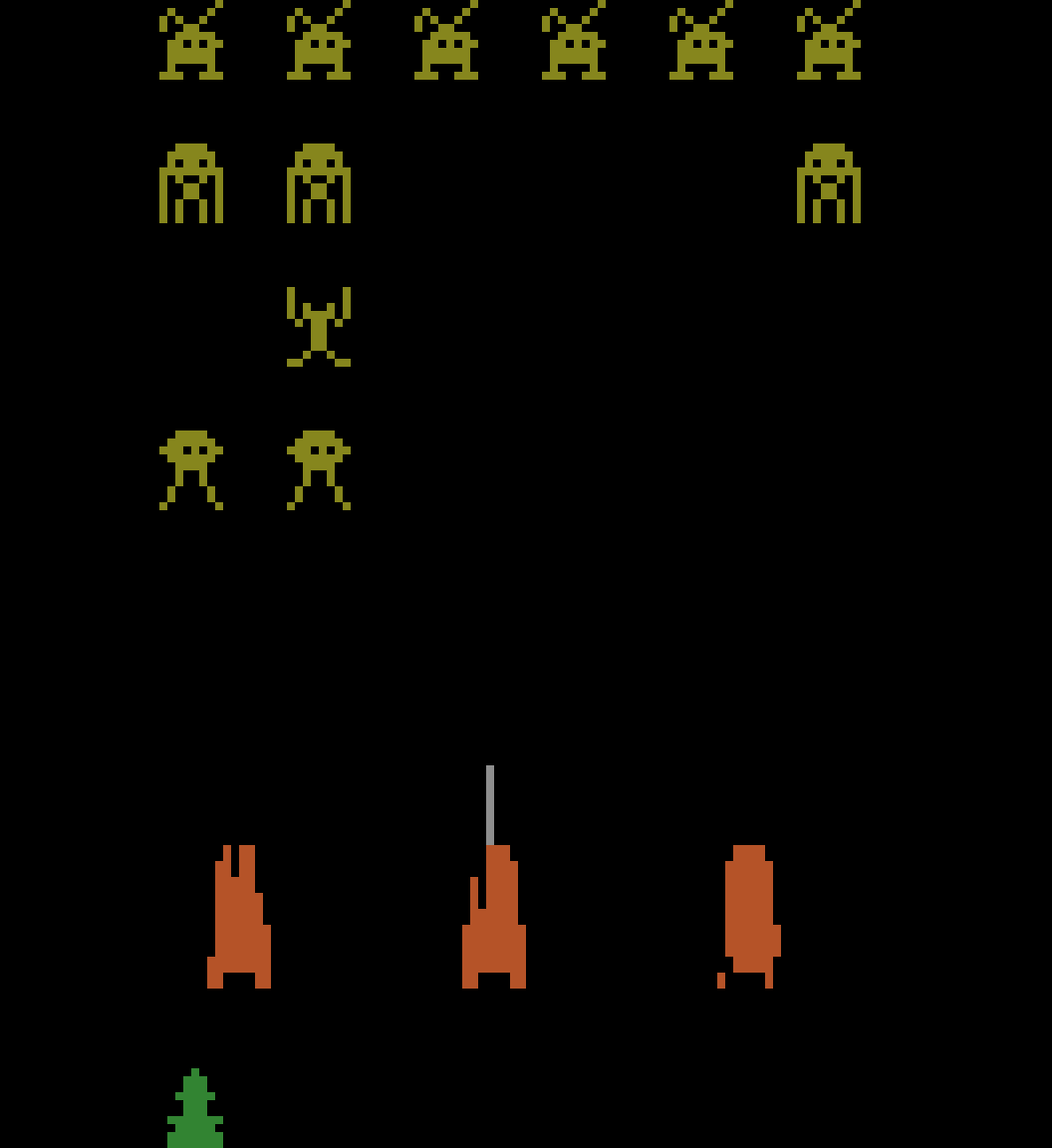} }}%
    \qquad
    {{\includegraphics[width=3.845cm]{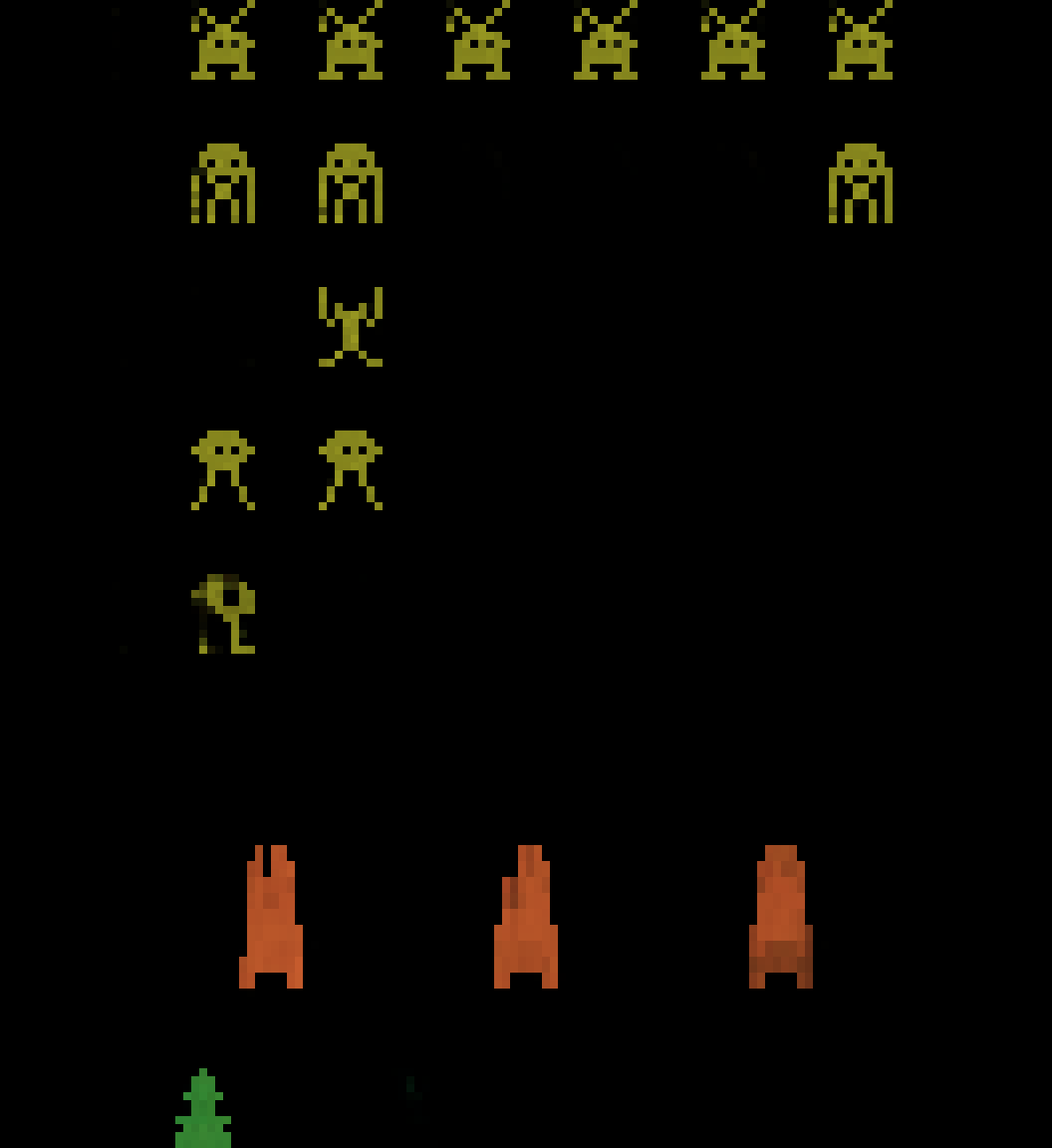} }}%
    \caption{\textbf{Left}: Space Invaders game state $\bm{s}$ in which an agent takes action $\bm{a} = \text{``move right''}$.
    \label{fig:example}%
    \textbf{Right}: counterfactual state $\bm{s'}$ where the agent will take the action $\bm{a'} = \text{``shoot''}$.}
\end{figure}

In this work, we introduce the concept of a ${\it counterfactual}$ ${\it state}$ as a counterfactual explanation. More precisely, for an agent in state $s$ performing action $a$ according to its learned policy, a counterfactual state $s'$ is a state that involves a minimal change to $s$ such that the agent's policy chooses action $a'$ instead of $a$. Figure \ref{fig:example} illustrates a counterfactual state for Space Invaders. Our approach is primarily intended for deep RL agents that operate in visual input environments, such as Atari. The main role of deep learning in these environments is to learn a low dimensional representation of the state to help with policy learning. Our approach investigates how changes to the state cause the agent to choose a different action. As such, we do not focus on explaining the long term, sequential decision making effects of following a learned policy, though this is a direction of interest for future work. 

Our end goal is a tool for acceptance testing for end users of a deep RL agent. We envision counterfactual states being used in a replay environment in which a human user observes the agent as it executes its learned policy. At key frames in the replay, the user can ask the agent to generate counterfactual states which help the user determine if the agent has captured relevant aspects of the visual input for its decision making.

The main contribution of this paper is a novel method to create counterfactual states through a deep generative architecture. Our approach can flexibly generate counterfactual states by moving through the deep network's latent space. We also investigate the realism and usefulness of these counterfactual states through a user study involving 30 participants who are not experts in machine learning. Our results show that our counterfactual states are realistic enough to improve participants' understanding of the agent's decision making.

%% file: relatedwork.tex
The literature on explainable AI is vast and we briefly summarize only the most directly related work.  Much of the past work on explaining machine learning has focused on explaining what features were important for a prediction (e.g. \cite{Ribeiro16}) or on identifying regions of the visual input that cause the agent to perform a certain action (e.g. \cite{pmlr-v80-greydanus18a,yang2018learn}).  These approaches do not use counterfactuals to explain machine learning and are thus orthogonal to our work. Other past work for explaining RL has looked at explaining policies through t-SNE embeddings \cite{Mnih15,Khan09}, state abstractions \cite{zahavy2016graying}, human-interpretable predicates \cite{hayes2017improving} and a high-level programming language \cite{Verma18}. These techniques look at explaining the policy trajectory which differs from our focus on how changes to the current visual input cause an alternate action to be chosen. Finally, \briefCite{huang2018} show that inspecting an RL agent's actions in critical states can improve end-user trust. 

Some recent techniques, not specific to RL, have looked at identifying differences that would cause an image be classified as another class. The Contrastive Explanations Method (CEM) \cite{Dhurandhar18} identifies critical features that must be present or absent for the predicted class. We apply CEM to our counterfactual task in Section \ref{sec:CEM} and discuss the issues. CEM has also been extended to explain differences between policies in reinforcement learning \cite{vanderWaa18}; this approach focuses on differences between trajectories and is different from our task. 

In recent work, methods to generate counterfactual explanations have also been applied in computer vision. Chang et al. \shortcite{chang2019explaining} generate counterfactuals by determining which regions, when filled in with values drawn from a "realistic" distribution, would most change the predicted class of the image.  Goyal et al. \shortcite{goyal:19} find the minimal number of region replacements that cause an image to be classified as a different class. When applied to our domain, these two counterfactual methods result in unrealistic counterfactuals because Atari images obey the rules of the game, which are easily violated with simple region replacements or in-filling. Our technical approach is also different as it requires following a gradient in latent space to generate the counterfactual image; moving about in latent space results in more flexibility for generating counterfactuals than infilling, but it introduces the risk of unrealistic images if the latent space is not well-behaved.



%% file: methodology.tex

\begin{figure}[t]
\begin{center}
\includegraphics[width=0.6\columnwidth]{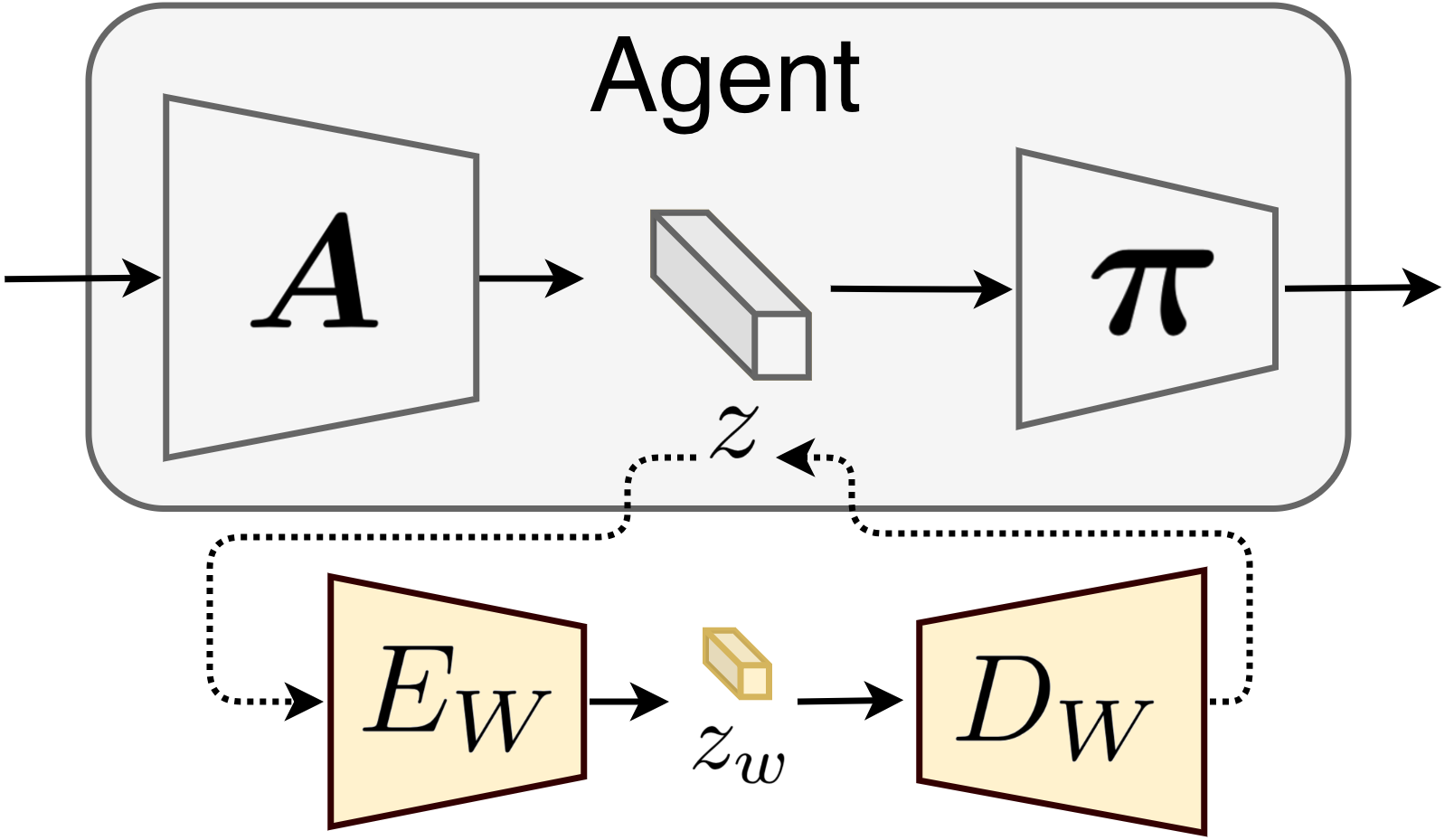}
\caption{The Wasserstein auto-encoder $E_W, D_W$ approximates the distribution of internal agent states $z$.}
\label{fig:train_architecture2}
\end{center}
\end{figure}

\begin{figure}[t]
\begin{center}
\includegraphics[width=0.75\columnwidth]{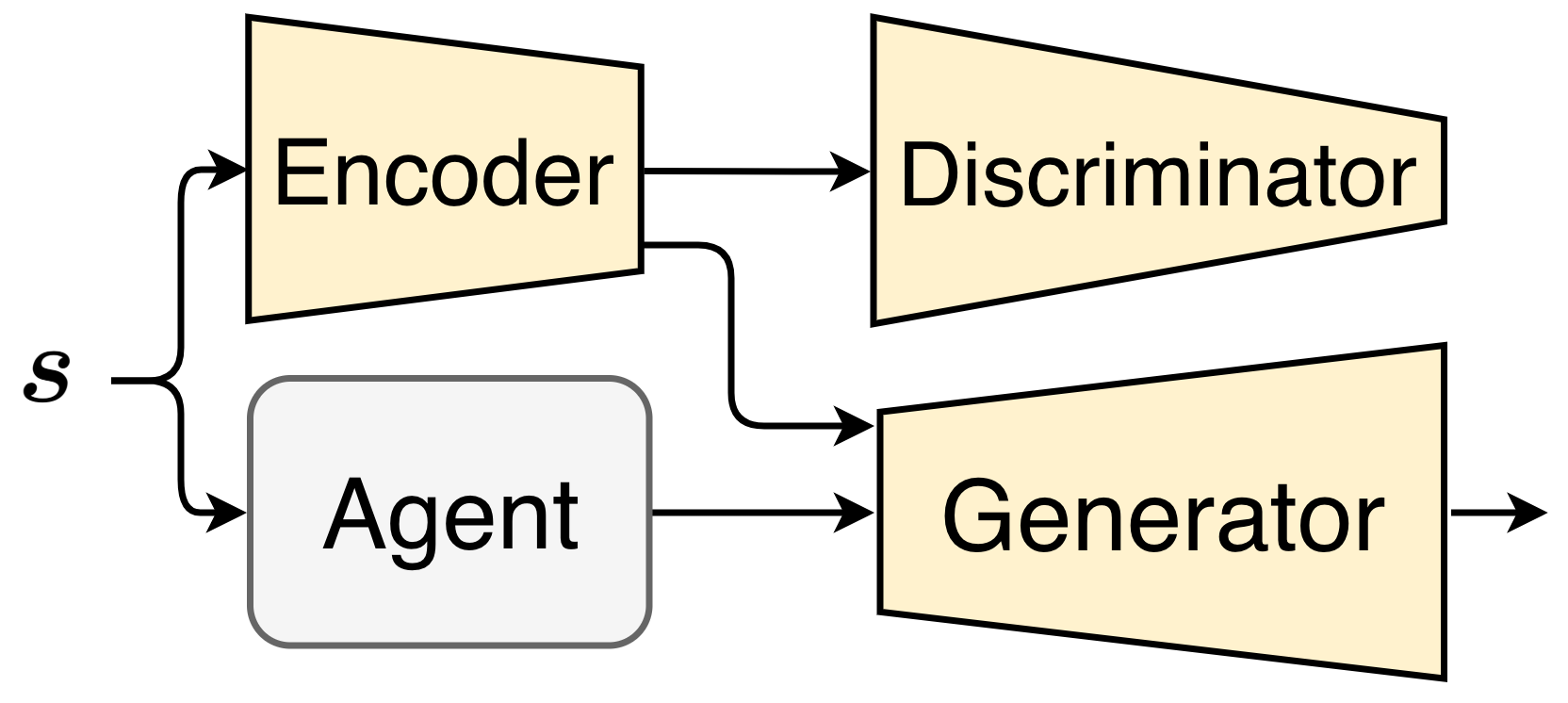}
\caption{The encoder $E$, generator $G$, and discriminator $D$ learn a model of the external environment for the pre-trained agent (grey).}
\label{fig:train_architecture1}
\end{center}
\end{figure}

The goal of this work is to shed some light into the decision making of a trained deep RL agent through counterfactual explanations. We are specifically interested in gaining some insight into what aspects of the visual input state $\bm{s}$ inform the choice of action $a$. Given a query state $\bm{s}$, we generate a \emph{counterfactual state} $\bm{s}'$ that minimally differs in some sense from $\bm{s}$, but results in the agent performing action $a'$ rather than action $a$. We refer to $a'$ as the \emph{counterfactual action}. 

Before providing an overview of our approach, we first introduce the notation we will use. As is typically done, vectors and matrices are boldfaced while scalars are not. Our approach requires a trained deep RL agent, which has a learned policy represented by a deep neural network. We divide this policy network into two partitions of interest (Figure \ref{fig:train_architecture2}). The first partition of the network layers, which we denote as $A$, takes a state $\bm{s}$ and maps it to a latent representation $\bm{z}=A(\bm{s})$. The vector $\bm{z}$ corresponds to the latent representation of $\bm{s}$ in the second to last fully connected layer in the network. The second partition of network layers, which we denote as $\bm{\pi}$, takes $\bm{z}$ and converts it to an action distribution $\bm{\pi}(\bm{z})$ i.e. a vector of probabilities for each action. Typically, $\bm{\pi}$ consists of a fully connected linear layer followed by a softmax. We use $\bm{\pi}(\bm{z},a)$ to refer to the probability of action $a$ in the action distribution $\bm{\pi}(\bm{z})$. In our Atari setting, it is important to distinguish between a state $\bm{s}$, which is a raw Atari game image (also called a game frame), and the latent state $\bm{z}$ which is obtained from the second to last fully connected layer of the policy network. This latent layer, which we call $\bm{Z}$ is important in our diagnosis because it is used by the agent to inform its choice of actions. In summary,
the agent\footnote{The agent may have other components such as the value function network. Our current work only uses the policy network, but we would like to apply similar ideas to the value function network.} can be viewed as the mapping $\bm{\pi}(A(\bm{s}))$.

In order to train our generative model, we require a training dataset $\mathcal{X} = \{(\bm{s}_1,\bm{a}_1),\ldots,(\bm{s}_N,\bm{a}_N)\}$ of $N$ state-action pairs. Here, the actions $\bm{a}_i$ are action distributions obtained from the trained agent as it executes its learned policy.


Our approach to counterfactual explanations is to create counterfactual states using a deep generative model, which have been shown to produce realistic images \cite{radford2015unsupervised}. Our primary strategy is to move in the latent space $Z$ in a direction that increases the probability of performing the counterfactual action $a'$.
However, as numerous researchers have noted, the latent space of a standard auto-encoder is filled with ``holes'' and counterfactual states generated from these holes would look unrealistic \cite{bengio2013representation}.
To produce a latent space that is more amenable to creating representative outputs, we create a novel architecture that involves an adversarial auto-encoder \cite{Makhzani15} and a Wasserstein auto-encoder \cite{tolstikhin2018wasserstein}.

\subsection{The Deep Network Architecture}

Figure \ref{fig:train_architecture1} depicts the architecture that we use during training. The RL agent is shaded gray to indicate that it has already been trained and is given to us. There are four components to this architecture: the Encoder ($E$), the Discriminator ($D$), the Generator ($G$) and the Wasserstein auto-encoder ($E_w, D_w$). Each of these components contributes a loss term to the overall loss function used to train the network. The subsections below will describe these components in turn.

\paragraph{Auto-encoder Loss}
The encoder $E$ and generator $G$ act as an encoder-decoder pair, with the task of creating reconstructed states when combined with the information from $\bm{\pi}(\bm{z})$. $E$ is a deep convolutional neural network that maps an input state $\bm{s}$ to a lower dimensional latent representation $E(\bm{s})$. $G$ is a deep convolutional generative neural network that creates an Atari image given its latent representation $E(\bm{s})$ and a policy vector $\bm{\pi}(\bm{s})$. The auto-encoding loss function of E and G is mean squared error (MSE): 

\begin{equation}
L_{AE} = \frac{1}{|\mathcal{X}|} \sum_{(\bm{s},\bm{a}) \in \mathcal{X}} ||G(E(\bm{s}),\bm{\pi}(A(\bm{s}))) - \bm{s} ||^2_2
\label{eqn:autoencoderloss}
\end{equation}

To generate counterfactual states, we want to create a new image by changing the action distribution $\bm{\pi}(A(\bm{s}))$ to reflect the desired counterfactual action $a'$.
However, the loss function $L_{AE}$ by itself will cause $G$ to ignore $\bm{\pi}(A(\bm{s}))$ and use only $E(\bm{s})$. 
We address this issue with an adversarial loss using a discriminator $D$.


\paragraph{Discriminator Loss} To ensure $\bm{\pi}(\bm{z})$ is not ignored, we cause the encoder to create an action-invariant representation $E(\bm{s})$.
By action-invariant, we mean that the representation $E(\bm{s})$ no longer captures aspects of the state $\bm{s}$ that inform the choice of action.
By doing so, adding $\bm{\pi}(\bm{z})$ as an input to $G$, along with $E(\bm{s})$, will provide the necessary information that will allow $G$ to recreate the effects of $\bm{\pi}$.
In order to create an action-invariant representation, we follow \cite{lample2017fader} and perform adversarial training on the latent space. 

We add a discriminator $D$ that is trained to predict the full action distribution $\bm{\pi}(\bm{z})$ given $E(\bm{s})$.
The action-invariant latent representation is learned by $E$ such that $D$ is unable to predict the true $\bm{\pi}(\bm{z})$ of our agent $A$.
As in Generative Adversarial Networks (GANs) \cite{goodfellow2014generative}, this setting corresponds to a two-player game where $D$ aims at maximizing its ability to identify the action distribution, and $E$ aims at preventing $D$ from being a good discriminator. 

The discriminator $D$ approximates $\bm{\pi}(\bm{z})$ given the encoded state $E(\bm{s})$, and is trained with MSE loss.

\begin{equation}
L_D =  \frac{1}{|\mathcal{X}|} \sum_{(\bm{s},\bm{a}) \in \mathcal{X}} ||D(E(\bm{s})) - \bm{\pi}(A(\bm{s})) ||^2_2
\label{eqn:discriminatorloss}
\end{equation}

\paragraph{Adversarial Loss} The objective of the encoder $E$ is now to learn a latent representation that optimizes two objectives.
The first objective causes the generator to reconstruct the state $\bm{s}$ given $E(\bm{s})$ and $\bm{\pi}(A(\bm{s}))$, but the second objective causes the discriminator to be unable to predict $\bm{\pi}(A(\bm{s}))$ given $E(\bm{s})$.
To accomplish this behavior in $D$, we want to maximize the entropy $H(D(E(\bm{s})))$, where $H(\bm{p}) = - \sum_i p_i log (p_i)$ Therefore, the adversarial loss can be written as:

\begin{equation}
L_{Adv} = \frac{\lambda}{|\mathcal{X}|} \sum_{(\bm{s},\bm{a}) \in \mathcal{X}} -H(D(E(\bm{s})))
\label{eqn:adversarialloss}
\end{equation}

In Equation \ref{eqn:adversarialloss}, $\lambda > 0$ weights the importance of this adversarial loss in the overall loss function.
A larger $\lambda$ amplifies the importance of a high entropy $\bm{\pi}(\bm{z})$, which in turn reduces the amount of action-related information in $E(\bm{s})$ and if pushed to the extreme, results in the Generator $G$ producing unrealistic game frames.
On the other hand, small values of $\lambda$ lower $G$'s reliance on the input $\bm{\pi}(\bm{z})$, resulting in small changes to the game state when $\bm{\pi}(\bm{z})$ is modified.

\paragraph{Wasserstein Autoencoder}


The counterfactual states require a notion of closeness between the query state $\bm{s}$ and the counterfactual state $\bm{s}'$. We can measure closeness in terms of distance in the agent's latent representation space $\bm{Z}$. We want to create a counterfactual state using $\bm{Z}$ as it directly influences the action distribution $\bm{\pi}$. We can perform gradient descent in this feature space with respect to our target action to produce a new $\bm{\pi}$ that has an increased probability of the counterfactual action $a'$. However, as previously mentioned, a latent representation may have holes in it \cite{bengio2013representation}, resulting in unrealistic counterfactuals. To avoid this problem, we re-represent $\bm{Z}$ to a lower-dimensional manifold $\bm{Z_W}$ that is more compact and better-behaved for producing realistic counterfactuals.

We use a Wasserstein auto-encoder (WAE) to learn a mapping function between the agent's feature space, to a well-behaved manifold \cite{tolstikhin2018wasserstein}. By using the concept of optimal transport, WAEs have shown they can learn not just a low dimensional embedding, but also one where data points retain their concept of closeness.

The closeness-preserving nature of the WAE plays an important role when creating an action distribution vector $\bm{\pi}(\bm{z})$.
In our counterfactual setting, we want to investigate the effect of performing action $a'$.
However, we cannot simply assign $a'$ a probability of 1 in the action distribution vector as this could result in unrepresentative/unrealistic images.
Instead, we follow a gradient in the $\bm{Z_W}$ space, which produces action distribution vectors that are more representative of those produced by the RL agent; this in turn results in more realistic images by the Generator $G$.

We train a WAE, with encoder $E_W$ and decoder $D_W$, on the agent's latent space $Z$ (see Figure~\ref{fig:train_architecture2}). We use MSE loss regularized by Maximum Mean Discrepancy (MMD):


\begin{align}
L_{\text{WAE}} =&
\frac{1}{|\mathcal{X}|} \sum_{\bm{s}} ||D_W(E_W(\bm{z})) - \bm{z} ||_2^2 +  \text{MMD}(D_W,E_W)
\label{eqn:wassersteinloss}
\end{align}
where $\bm{z} = A(\bm{s})$ and the $\text{MMD}$ is between $D_W$ and $E_W$ in $\mathbf{Z}$ measured using an inverse multiquadratic kernel \cite{tolstikhin2018wasserstein}.




\paragraph{Training}
We let the previously trained agent play the game with $\epsilon$-greedy exploration and train with the resulting dataset $\mathcal{X} = \{(\bm{s}_1, \bm{a}_1), \ldots, (\bm{s}_N,\bm{a}_N))$. 
We train: the Encoder and Generator to minimize reconstruction error (Equation \ref{eqn:autoencoderloss}), the discriminator to predict the action probabilities (Equation \ref{eqn:discriminatorloss}), the encoder to adversarially fool the discriminator (Equation \ref{eqn:adversarialloss}), and the WAE to minimize both the reconstruction of agent's latent state representation as well as the MMD (Equation \ref{eqn:wassersteinloss}). This is done at each game time step with stochastic gradient descent \cite{ADAM}




\subsection{Generating Counterfactuals} 
\label{sec:generating_counterfactuals}
Our goal is to use counterfactual image generation to create synthetic images that closely resemble real states of the game environment, but result in the agent taking action $a'$ instead of action $a$.
Similar to \cite{neal2018open}, we formulate this as an optimization:

\begin{align*}
    \text{minimize} &&
    ||E_w(A(\bm{s})) - \bm{z_w^*}||_2^2 \\
    \text{subject to} &&
    \argmax_{a \in \text{\textit{Actions}}}\ \  \bm{\pi}(D_W(\bm{z_w^*}), a) = a'
\end{align*}

where $\bm{s}$ is the given query state and $\bm{z_w^*}$ is a latent point representing a possible internal state of the agent.
This optimization can be relaxed as follows:

\begin{equation} \label{grad_descent}
\bm{z_w^*} = \argmin_{\bm{z_w}}
||\bm{z_w} - E_w(A(\bm{s}))||_2^2
+
\log \left(1 - \bm{\pi}(D_W(\bm{z_w}), a' ) \right)
\end{equation}

where $\bm{\pi}(\bm{z}, a)$ is the probability of the agent taking a discrete action $a$ on the counterfactual state representation $\bm{z}$.
Minimizing the second term increases the probability of taking action $a'$ and reduces the probability of taking all other actions. 

We generate a counterfactual state by selecting a state $\bm{s}$ from the training set, then encoding the state to a Wasserstein latent point $\bm{z_w} = E_W(A(\bm{s}))$.
We then minimize Equation \ref{grad_descent} with gradient descent to find $\bm{z_w^*}$. The latent point $\bm{z_w^*}$ is decoded to create  $\bm{\pi}(D_w(\bm{z_w^*)})$ which is passed to the generator, along with $E(\bm{s})$ to create the counterfactual state $\bm{s'}$.





%% file: experiments.tex
%

\subsection{Experimental Setup}
The pre-trained agent is a deep convolutional network trained with Asynchronous Advantage Actor-Critic (A3C) to maximize score in an Atari game. Games are played with a fixed frame-skip of 8 (7 for Space Invaders). We decompose the agent into two functions: $A(\bm{s})$ which takes as input 4 concatenated video frames and produces a 256-dimensional vector $\bm{z}$, and $\bm{\pi}(z)$ which outputs a distribution among actions. To generate the dataset $\mathcal{X}$, we set $\epsilon$ exploration value to $0.2$ and have the agent play for 40 million environment steps.  

The encoder $E$ consists of 6 convolutional layers followed by 2 fully-connected layers with LeakyReLU activations and batch normalization. The output $E(\bm{s})$ is a 16-dimensional vector. We find a value of $\lambda=20$ enforces a good tradeoff between state reconstruction and reliance on $\bm{\pi}(z)$. 
The generator $G$ consists of one fully-connected layer followed by 6 transposed convolutional layers, all with ReLU activations and batch normalization.
The encoded state $E(\bm{s})$ and the action distribution $\bm{\pi}(\bm{z})$ are fed to the first layer of the generator, and additionally $\bm{\pi}(\bm{z})$ is appended as an additional input channel to each subsequent layer.
The discriminator $D$ consists of two fully-connected layers followed by a softmax function, and outputs a distribution among actions with the same dimensionality as $\bm{\pi}(\bm{z})$.
The Wasserstein encoder $E_w$ consists of 3 fully-connected layers mapping $\bm{z}$ to a 128-dimensional vector $\bm{z}_w$.
The corresponding Wasserstein decoder $D_w$ is symmetric to $E_w$ and maps $\bm{z}_w$ back to $\bm{z}$.

All models are constructed and trained using PyTorch \cite{paszke2017automatic}.


\subsection{Baseline Comparisons}

\begin{figure}[t]
    \centering
    \includegraphics[width=.32\linewidth]{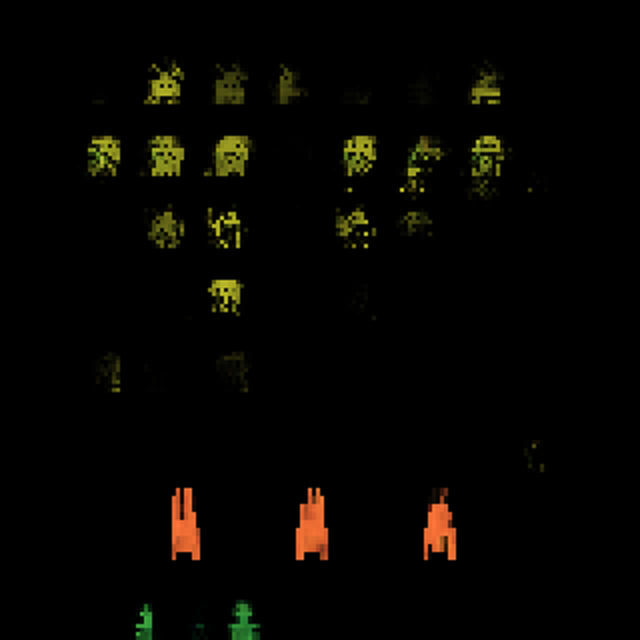} 
    \includegraphics[width=.32\linewidth]{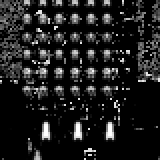} 
    \includegraphics[width=.32\linewidth]{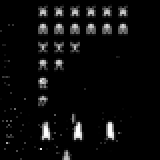} %
    \caption{Counterfactual states generated using the ablated model (\textbf{Left}), and CEM with two choices of parameters on different states (\textbf{Middle} and \textbf{Right}).}
    
    \label{fig:CEM}%
\end{figure}

We compare our full method against an ablated version trained without adversarial loss, and against the Contrastive Explanation Method (CEM) \cite{Dhurandhar18}.

In the ablated version of the network, the encoder, discriminator, and Wasserstein autoencoder are removed, and the generator is trained with MSE loss to reconstruct $\bm{s}$ given $\bm{z}$ as input.
Counterfactual images are generated by performing gradient descent with respect to $\bm{z}$ to maximize $\bm{\pi}(\bm{z},a')$ for a counterfactual action $a'$.
We find that counterfactual states generated in this way fail to construct a fully realistic state as shown in Figure \ref{fig:CEM} (left).


\label{sec:CEM}
The Contrastive Explanation Method (CEM) can generate \textit{pertinent negatives} which highlight absent features would cause the agent to select an alternate action.
We generate pertinent negatives from Atari states with pixels as features, and interpret them as counterfactual states.
We performed an extensive search over hyper-parameters to generate realistic states, but found CEM difficult to tune for this high-dimensional space. The generated counterfactual states were either identical to the original query state or they had excessive ``snow'' artifacts as shown in Figure \ref{fig:CEM} (middle and right).


\subsection{Example Counterfactual States}


\begin{figure*}[t]
    \centering
 \includegraphics[width=\linewidth]{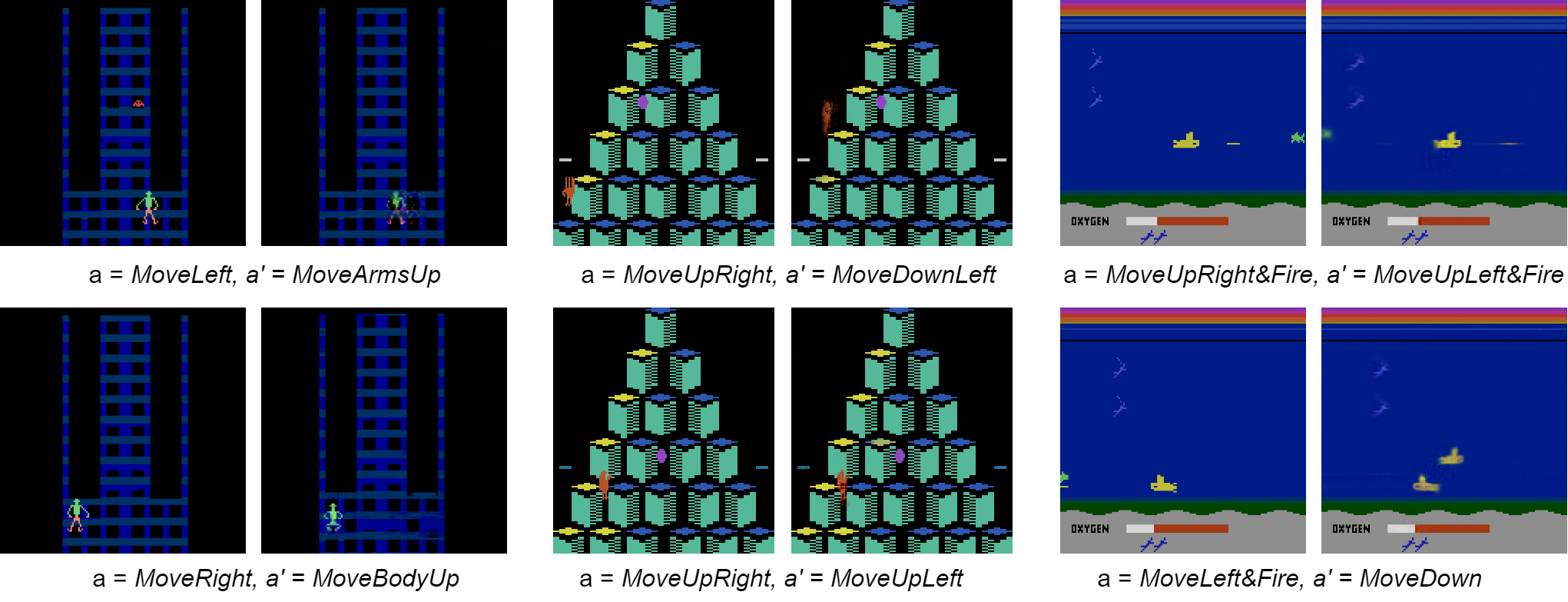}
    \caption{
    \textbf{Left Column}: Crazy Climber, \textbf{Center Column}: Q*bert, \textbf{Right Column}: Seaquest.
    Paired examples of query state with action $a$ (left) and counterfactual state with action $a'$ (right).
     }
    \label{fig:cf_imgs}%
\end{figure*}

\begin{figure*}[h]
\includegraphics[width=.98\linewidth]{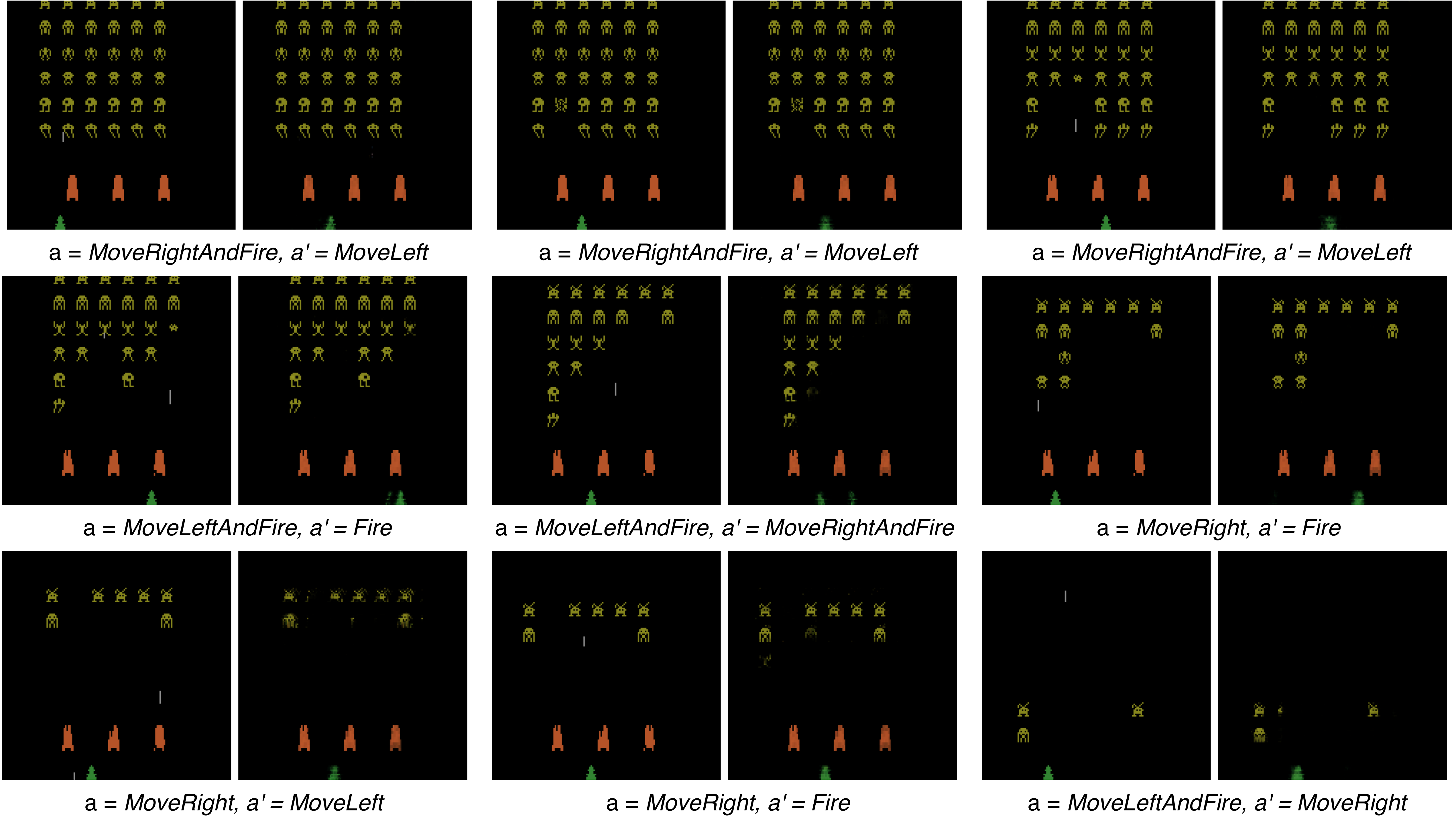}
    \caption{User study images of Space Invaders, with pairs of images showing the query state with action $a$ (left) and the counterfactual state with action $a'$ (right).}
    \label{fig:space_images}
\end{figure*}

We now show examples of counterfactual states for pre-trained agents in various Atari games. Figure \ref{fig:cf_imgs} and \ref{fig:space_images} demonstrates pairs of images where the 
the left image is the original query state where the agent would take action $a$ according to its policy, and the right image is the counterfactual state where the agent would take the selected action $a'$. 

In {\em Crazy Climber}, an agent must climb up a building while avoiding various obstacles.  Figure \ref{fig:cf_imgs} (Top-Left) shows how the agent will climb up as the enemy is no longer above it. In Figure \ref{fig:cf_imgs} (Bottom-Left), the original state shows the agent in a position to move horizontally, where the counterfactual state shows the climber in a ready state to move vertically.




In {\em $\text{Q}^*$bert}, an agent's goal is to jump on uncolored squares and avoid enemies. In Figure \ref{fig:cf_imgs} (Top-Center), the agent will jump down and left if the Qbert character had been higher up on the structure. In Figure \ref{fig:cf_imgs} (Bottom-Center), the counterfactual shows that the up-right square is yellow (already visited), which will cause Qbert to move up-left.


%
%

In {\em Seaquest}, an agent must shoot incoming enemies while rescuing friendly divers.  Figure \ref{fig:cf_imgs} (top-right) shows that a new enemy must appear to the left in order for the agent to take an action that turns the submarine around while firing. Thus, the agent has an understanding about enemy spawns and submarine direction.
Figure \ref{fig:cf_imgs} (Bottom-Right) shows an unrealistic counterfactual with two submarines.

In {\em Space Invaders}, an agent exchanges fire with approaching enemies while taking cover underneath three barriers. The examples in Figure \ref{fig:space_images} reveal the agent has learned to prefer specific locations for safely lining up shots.




\subsection{User Study}
Evaluating counterfactual states is a challenging problem. Good counterfactual states provide insight to a human as to why the agent performed a certain action. This criterion is difficult to capture through quantitative metrics, which often measure the wrong thing. For instance, using the probability $\pi(\bm{s'},a')$ as a metric for a counterfactual state $\bm{s}'$ is misleading because this probability can be swayed by Atari images that are obvious unrealistic to humans or by adversarial examples with imperceptible changes to the original state $\bm{s}$. 

As a result, we evaluated our counterfactual states through a user study in our lab with participants who were not experts in machine learning; participants included undergraduates and members of the local community. We chose to focus our study on Space Invaders because it is straightforward to learn for a participant unfamiliar with video games. For this reason, we started the study by having participants play Space Invaders for 5 minutes. Participants then rated the realism of 30 randomly ordered game images on a Likert scale from 1 to 6 (higher is more realistic). The images were a mix of images selected from three different sources: the actual game, our counterfactual method, and our ablation experiment, with 10 images from each source. 

For the next part of the study, we had participants watch a replay of the agent playing a game of space invaders; following this, the participants were given a tutorial to explain counterfactual states. We found that a guided in-person tutorial was helpful to clarify participant confusion about counterfactuals, which was an unfamiliar topic to many participants.

We then showed the users 10 counterfactual states displayed alongside the original query state and an image that highlights the changes. We chose the game images from a replay of an existing game. The specific images, serving as query states for our counterfactuals, were chosen using a heuristic based on entropy, which has been used in the past for choosing key frames for establishing trust \cite{huang2018}. 
For diversity, if a key frame was selected, we do not allow images from the next two time-steps to be selected. 
Once a query state was selected, we selected the counterfactual action $a'$ as the one that required the largest $L_2$ change between the original Wasserstein latent state $\bm{z_w}$ and the counterfactual one $\bm{z_w'}$ (ignoring the no-operation action). 
The set of images generated by this method and used in the study are shown in Figure \ref{fig:example} and \ref{fig:space_images}. We emphasize that the counterfactual states were not hand-picked; rather, they were selected by ranking game images according to our heuristic and then selecting the counterfactual action according to the previous criterion.
 
\paragraph{Study Results} In terms of realism, the average ratings on the 6 point Likert scale were 1.93 (ablation), 4.00 (counterfactual states) and 4.97 (actual game). The differences between the realism ratings for the counterfactual states and real states were not statistically significant ($\alpha=0.05$, p-value=0.458, one-sided Wilcoxon signed-rank test). These results show that our counterfactual states were on average close to appearing realistic, but there were some flaws. 

We also asked the participants to rate their understanding of the agent on a 6 point Likert scale before and after seeing the counterfactual states. We found 15 users' understanding increased, 8 decreased, and 7 stayed the same (p-value = 0.098, one-sided Wilcoxon signed-rank test).  These results are close to being statistical significant at the $\alpha=0.05$ level and they suggest that counterfactual states are indeed providing most users with enough insight into an agent's decision making to improve their understanding of how they work.

%% file: discussion.tex

We end with a brief discussion of some important issues with our approach. 
First, our deep generative approach adds some artifacts when creating counterfactual states, which impacts the faithfulness of our explanation. Empirically, we found most artifacts were fairly minor, such as blurry images, and did not seem to be a major roadblock for our participants. One of the more noticeable artifacts is how small objects, such as the shot in space invaders, disappear. These small objects, however, could be important for other domains (e.g. Pong). It is likely that some of these artifacts could be fixed by training longer, with more data, and with better architectures. This problem also raises an open question in representation learning about preserving small, but important, objects in images. 
A second issue is how to select query states from a replay such that the counterfactual states, and actions, provide the most insight to a human. Our criterion was based on heuristics and a deeper investigation is needed. 
A third issue is with the evaluation for understanding. We acknowledge an objective metric is preferred for evaluating understanding. A first thought, for measuring the effectiveness of counterfactual explanations, would be having participants predict an agent's action in a new state-- but \cite{anderson2019mortals} show that utilizing explanations, to try and predict future actions, is difficult as an agent's choice can appear to be counter-intuitive. Though with more research, it is likely that a suitable task to evaluate counterfactual explanations can be found.

%% file: conclusion.tex
We introduced a deep generative model to produce counterfactual states to provide insight into a deep RL agent's decision making. The counterfactual states show what minimal changes need to occur to a state to produce a different action by the trained RL agent. Our results indicate these counterfactual states are fairly realistic, but do contain some artifacts. 


For future work, we will investigate how counterfactuals extend to domains beyond Atari games. In addition, we plan to apply counterfactual explanations to the problem of explaining long range sequential decision making aspects in RL. 


%% file: main.bbl
\begin{thebibliography}{}

\bibitem[\protect\citeauthoryear{Anderson \bgroup \em et al.\egroup
  }{2019}]{anderson2019mortals}
Andrew Anderson, Jonathan Dodge, Amrita Sadarangani, Zoe Juozapaitis, Evan
  Newman, Jed Irvine, Souti Chattopadhyay, Alan Fern, and Margaret Burnett.
\newblock Explaining reinforcement learning to mere mortals: An empirical
  study.
\newblock In {\em In Proceedings of the 28th International Joint Conference on
  Artificial Intelligence}, 2019.

\bibitem[\protect\citeauthoryear{Bengio \bgroup \em et al.\egroup
  }{2013}]{bengio2013representation}
Yoshua Bengio, Aaron Courville, and Pascal Vincent.
\newblock Representation learning: A review and new perspectives.
\newblock {\em IEEE transactions on pattern analysis and machine intelligence},
  35(8):1798--1828, 2013.

\bibitem[\protect\citeauthoryear{Chang \bgroup \em et al.\egroup
  }{2019}]{chang2019explaining}
Chun-Hao Chang, Elliot Creager, Anna Goldenberg, and David Duvenaud.
\newblock Explaining image classifiers by counterfactual generation, 2019.
\newblock To appear at the Seventh International Conference on Learning
  Representations.

\bibitem[\protect\citeauthoryear{Dhurandhar \bgroup \em et al.\egroup
  }{2018}]{Dhurandhar18}
Amit Dhurandhar, Pin-Yu Chen, Ronny Luss, Chun-Chen Tu, Paishun Ting,
  Karthikeyan Shanmugam, and Payel Das.
\newblock Explanations based on the missing: Towards contrastive explanations
  with pertinent negatives.
\newblock In S.~Bengio, H.~Wallach, H.~Larochelle, K.~Grauman, N.~Cesa-Bianchi,
  and R.~Garnett, editors, {\em Advances in Neural Information Processing
  Systems 31}, pages 592--603. Curran Associates, Inc., 2018.

\bibitem[\protect\citeauthoryear{Goodfellow \bgroup \em et al.\egroup
  }{2014}]{goodfellow2014generative}
Ian Goodfellow, Jean Pouget-Abadie, Mehdi Mirza, Bing Xu, David Warde-Farley,
  Sherjil Ozair, Aaron Courville, and Yoshua Bengio.
\newblock Generative adversarial nets.
\newblock In {\em Advances in neural information processing systems}, pages
  2672--2680, 2014.

\bibitem[\protect\citeauthoryear{Goyal \bgroup \em et al.\egroup
  }{2019}]{goyal:19}
Yash Goyal, Ziyan Wu, Jan Ernst, Dhruv Batra, Devi Parikh, and Stefan Lee.
\newblock Counterfactual visual explanations.
\newblock In {\em Proceedings of the Thirty-Sixth International Conference on
  Machine Learning (ICML)}, 2019.

\bibitem[\protect\citeauthoryear{Greydanus \bgroup \em et al.\egroup
  }{2018}]{pmlr-v80-greydanus18a}
Samuel Greydanus, Anurag Koul, Jonathan Dodge, and Alan Fern.
\newblock Visualizing and understanding {A}tari agents.
\newblock In {\em Proceedings of the 35th International Conference on Machine
  Learning}, Proceedings of Machine Learning Research, 2018.

\bibitem[\protect\citeauthoryear{Hayes and Shah}{2017}]{hayes2017improving}
Bradley Hayes and Julie~A Shah.
\newblock Improving robot controller transparency through autonomous policy
  explanation.
\newblock In {\em Proceedings of the 2017 ACM/IEEE international conference on
  human-robot interaction}, pages 303--312. ACM, 2017.

\bibitem[\protect\citeauthoryear{Huang \bgroup \em et al.\egroup
  }{2018}]{huang2018}
Sandy~H. Huang, Kush Bhatia, Pieter Abbeel, and Anca~D. Dragan.
\newblock Establishing (appropriate) trust via critical states.
\newblock In {\em HRI 2018 Workshop: Explainable Robotic Systems}, 2018.

\bibitem[\protect\citeauthoryear{Jasper van~der Waa}{2018}]{vanderWaa18}
Karel van den Bosch Mark~Neerincx Jasper van~der Waa, Jurriaan van~Diggelen.
\newblock Contrastive explanations for reinforcement learning in terms of
  expected consequences.
\newblock In {\em Proceedings of the IJCAI/ECAI 2018 Workshop on Explainable
  AI}, pages 165--171, 2018.

\bibitem[\protect\citeauthoryear{Khan \bgroup \em et al.\egroup
  }{2009}]{Khan09}
Omar~Zia Khan, Pascal Poupart, and James~P. Black.
\newblock Minimal sufficient explanations for factored markov decision
  processes.
\newblock In {\em Proceedings of the Nineteenth International Conference on
  Automated Planning and Scheduling}, pages 194--200, 2009.

\bibitem[\protect\citeauthoryear{Kingma and Ba}{2014}]{ADAM}
Diederik~P. Kingma and Jimmy Ba.
\newblock Adam: {A} method for stochastic optimization.
\newblock {\em CoRR}, abs/1412.6980, 2014.

\bibitem[\protect\citeauthoryear{Lample \bgroup \em et al.\egroup
  }{2017}]{lample2017fader}
Guillaume Lample, Neil Zeghidour, Nicolas Usunier, Antoine Bordes, Ludovic
  Denoyer, et~al.
\newblock Fader networks: Manipulating images by sliding attributes.
\newblock In {\em Advances in Neural Information Processing Systems}, pages
  5967--5976, 2017.

\bibitem[\protect\citeauthoryear{Lewis}{1973}]{lewis2013counterfactuals}
David Lewis.
\newblock {\em Counterfactuals}.
\newblock John Wiley \& Sons, 1973.

\bibitem[\protect\citeauthoryear{Makhzani \bgroup \em et al.\egroup
  }{2015}]{Makhzani15}
Alireza Makhzani, Jonathon Shlens, Navdeep Jaitly, and Ian~J. Goodfellow.
\newblock Adversarial autoencoders.
\newblock {\em CoRR}, abs/1511.05644, 2015.

\bibitem[\protect\citeauthoryear{Mnih and Hassabis}{2015}]{Mnih15}
Kavukcuoglu K. Silver D. Rusu A. A. Ve-ness J. Bellemare M. G. Graves A.
  Riedmiller M. Fidjeland A. K. Ostrovski G. Petersen S. Beattie C. Sadik A.
  Antonoglou I. King H. Kumaran D. Wier-stra D. Legg~S. Mnih, V. and
  D.~Hassabis.
\newblock Human-level control through deep reinforcement learning.
\newblock {\em Nature}, 518(7540):529–533, 2015.

\bibitem[\protect\citeauthoryear{Neal \bgroup \em et al.\egroup
  }{2018}]{neal2018open}
Lawrence Neal, Matthew Olson, Xiaoli Fern, Weng-Keen Wong, and Fuxin Li.
\newblock {\em Open Set Learning with Counterfactual Images: 15th European
  Conference, Munich, Germany, September 8–14, 2018, Proceedings, Part VI},
  pages 620--635.
\newblock 09 2018.

\bibitem[\protect\citeauthoryear{Paszke \bgroup \em et al.\egroup
  }{2017}]{paszke2017automatic}
Adam Paszke, Sam Gross, Soumith Chintala, Gregory Chanan, Edward Yang, Zachary
  DeVito, Zeming Lin, Alban Desmaison, Luca Antiga, and Adam Lerer.
\newblock Automatic differentiation in pytorch.
\newblock 2017.

\bibitem[\protect\citeauthoryear{Radford \bgroup \em et al.\egroup
  }{2015}]{radford2015unsupervised}
Alec Radford, Luke Metz, and Soumith Chintala.
\newblock Unsupervised representation learning with deep convolutional
  generative adversarial networks.
\newblock {\em arXiv preprint arXiv:1511.06434}, 2015.

\bibitem[\protect\citeauthoryear{Ribeiro \bgroup \em et al.\egroup
  }{2016}]{Ribeiro16}
Marco~Tulio Ribeiro, Sameer Singh, and Carlos Guestrin.
\newblock "why should i trust you?": Explaining the predictions of any
  classifier.
\newblock In {\em Proceedings of the 22Nd ACM SIGKDD International Conference
  on Knowledge Discovery and Data Mining}, pages 1135--1144, New York, NY, USA,
  2016. ACM.

\bibitem[\protect\citeauthoryear{Tolstikhin \bgroup \em et al.\egroup
  }{2018}]{tolstikhin2018wasserstein}
Ilya Tolstikhin, Olivier Bousquet, Sylvain Gelly, and Bernhard Scholkopf.
\newblock Wasserstein auto-encoders.
\newblock 2018.

\bibitem[\protect\citeauthoryear{Verma \bgroup \em et al.\egroup
  }{2018}]{Verma18}
Abhinav Verma, Vijayaraghavan Murali, Rishabh Singh, Pushmeet Kohli, and Swarat
  Chaudhuri.
\newblock Programmatically interpretable reinforcement learning.
\newblock {\em CoRR}, abs/1804.02477, 2018.

\bibitem[\protect\citeauthoryear{Yang \bgroup \em et al.\egroup
  }{2018}]{yang2018learn}
Zhao Yang, Song Bai, Li~Zhang, and Philip~HS Torr.
\newblock Learn to interpret atari agents.
\newblock {\em arXiv preprint arXiv:1812.11276}, 2018.

\bibitem[\protect\citeauthoryear{Zahavy \bgroup \em et al.\egroup
  }{2016}]{zahavy2016graying}
Tom Zahavy, Nir Ben-Zrihem, and Shie Mannor.
\newblock Graying the black box: Understanding dqns.
\newblock In {\em International Conference on Machine Learning}, pages
  1899--1908, 2016.

\end{thebibliography}
